# Creative Action at a Distance: A Conceptual Framework for Embodied Performance With Robotic Actors

Philipp Wicke* and Tony Veale

*School of Computer Science, University College Dublin, Belfield, Ireland*

Acting, stand-up and dancing are creative, embodied performances that nonetheless follow a script. Unless experimental or improvised, the performers draw their movements from much the same stock of embodied schemas. A slavish following of the script leaves no room for creativity, but active interpretation of the script does. It is the choices one makes, of words and actions, that make a performance creative. In this theory and hypothesis article, we present a framework for performance and interpretation within robotic storytelling. The performance framework is built upon movement theory, and defines a taxonomy of basic schematic movements and the most important gesture types. For the interpretation framework, we hypothesise that emotionally-grounded choices can inform acts of metaphor and blending, to elevate a scripted performance into a creative one. Theory and hypothesis are each grounded in empirical research, and aim to provide resources for other robotic studies of the creative use of movement and gestures.

## 1 INTRODUCTION

Embodied performances on a stage often start with a script. Performers can slavishly follow this script, like a computer executing a computer program, or they can interpret its directives as they see fit. Only by doing the latter can a performer be said to deliver a "creative" performance.

A performance is a conceptual scheme turned into physical action. When concepts become movements, movements suggest meanings and meanings evoke concepts in the minds of an audience. Since every link in this chain is under-determined, creativity can insinuate itself into every part of the meaning-making process. The physical actions of a performance suggest meanings, or reinforce what is also communicated with words, so the most effective actions tap into an audience's sense of familiarity, obviousness and conceptual metaphor. In this paper, we consider story-telling as a performance that combines linguistic (spoken) and physical (embodied) actions. Our embodied actors can communicate a tale by narrating it, or by acting it out, or as an ensemble of agents that do both. Our focus is unique for a number of reasons. First, our embodied actors are robots, not humans, although they aim to move, pose and gesture much as humans do. Second, the tales they tell are not spun by a human, but generated by a machine in an act of computational creativity. An AI system that controls the writing process, the telling process and the acting process can thus be used to explore the ties between concepts, words and embodied actions in a creative, performative setting. Third, our robots can interpret the written script just as actors interpret a film script. They can literally depict the actions through movement and gesture, or they can interpret the actions of the script metaphorically. This flexible reading of the script allows metaphor to shape its embodied interpretation, fostering creativity in the physical enactment of the story. In short, we explore here how interpretation is infused with emotionally-grounded choice to appreciate and to achieve embodied creativity in a system for performing machine-generated stories.

Story-telling is just one kind of embodied performance. We humans use our bodies to tell jokes, engage in animated conversation, and communicate feelings in play and in dance. Starting from a narrative perspective, with a system designed to support the performance of computer-generated stories with computer-controlled robotic actors, we set out to generalize our approach and create a framework for embodied communication that can support multiple types of performance. Key to this approach is the meaning-making potential of physical acts, which we ground in image-schematic models of language. As we will show here, story-telling provides an ideal basis for empirically testing our hypotheses, but our aim is to broaden the framework to accommodate new possibilities and new kinds of performance.

We adopt a bottom-up approach to unifying theory and practice, in which an implemented AI system supports the empirical studies that motivate our hypotheses, before we generalize those hypotheses into a combinatorial framework for embodied meaning-making. We begin by surveying the state-of-the-art in robotic performance to define a taxonomy that accommodates humanoid movements from walking to posing and gesturing. Although physical actions are not words, deliberate physical actions do have a semiotic component that we will analyze here. So, by exploring robotic enactment in a storytelling context, we can identify the semantic units of movement and their cognitive-linguistic underpinnings in image schemas and conceptual metaphors. Ultimately, our goal is to identify the points of contact between action and meaning where creativity – and in particular, machine creativity – can blossom.

In our *Performance Framework* (**Section 3**), we outline which movements can be executed in parallel or in series, to convey meanings of their own or to augment the meaning of the spoken dialogue. In addition, we will consider the properties of physical actions to identify those that are additive (when compounded movements achieve a cumulative effect), persistent (when a movement has a lasting effect on the physical relationship between actors), and summative (allowing an action to summarize what has already occurred). For example, the meaning conveyed by one actor stepping away from another intensifies with each additional step. The action and its meaning are also persistent, since, unlike gestures, stepping away does not necessitate a subsequent retraction of the action. After multiple steps, the resulting distance between actors (and characters) is the sum of all steps, and so conveys a global perspective.

We distinguish between locomotive, spatial movements (hereinafter spatial movement) along a stage, postural reorganizations of the body, and gestures made with the hands, arms and upper-body to communicate specific intents. For gestures, we also discriminate pantomimic or iconic gestures (which play-act a meaning, e.g., *using an invisible steering wheel to signify driving*) from more arbitrary actions (which may use metonymy to depict culturally-specific actions, such as *bending the knee to propose*), from those that instantiate a conceptual metaphor to achieve their communicative intent. The framework formally integrates each of these forms of physical meaning-making, and constrains how they work with each other in the realization of a coherent performance.

Connecting the underlying script with the performance requires an appropriate choice of the movements to be enacted. When a script dictates the actions, there is no space for choice. Likewise, when a script provides simple disjunctive choices – do *this* or *that* – it allows a performer to explore the space of possible stories without regard for the emotions of the characters. The actor's capacity to interpret the script should consider these emotions and how they can shape the performance. By creating choices at the time of performance, an interpretation can look at the unfolding narrative so far and shape the course that the actor will take. Since these choices are guided by the character's emotional valence at any given moment, we introduce an emotional layer between the level of scripted actions (what happens next) and the expressive level (physical movements and spoken words). This new layer annotates the emotions implied for each action and movement, to inform the actors about the emotional resonances of their choices. By considering the influence of earlier actions in the plot, choices can be made in the moment, to reflect an interpretation of how characters should be feeling and acting. The *Interpretation Framework* (**Section 4**) provides the tools to a performer to make deliberate use of gesture and space for an emotionally-informed performance.

Since the AI system automatically maps tales from pre-generated texts onto physical performances, we can use these performances as the basis of empirical studies that explore whether audiences intuitively appreciate the deliberate use of space and gesture in a performance. More specifically, we look at whether coherent usage is as appreciated as incoherent use, and whether the schematic use of space in a cumulative, summative fashion is as appreciated by audiences as the use of transient, culturally-specific gestures. We interpret the results of those studies with respect to the frameworks presented here. Both studies (described in **Section 5**) have been conducted by recording robotic performances under the coherent and incoherent conditions, and then eliciting crowd-sourced ratings of those performances. Each participant is shown short videos of plot segments that feature relevant movements, and each is asked to rate the performance on a customized HRI questionnaire. As such, we intend to contribute to multiple areas of interdisciplinary research with this framework: not just automated storytelling (as built on automated story-generation) and embodied performance using robots, but the study of expressive gestures and physical meaning-making more generally, across a diversity of settings. While we evaluate a rather specific use of the framework in the story domain, we will provide a taxonomy and a terminology that will foster further interdisciplinary research in the areas that contribute to it.

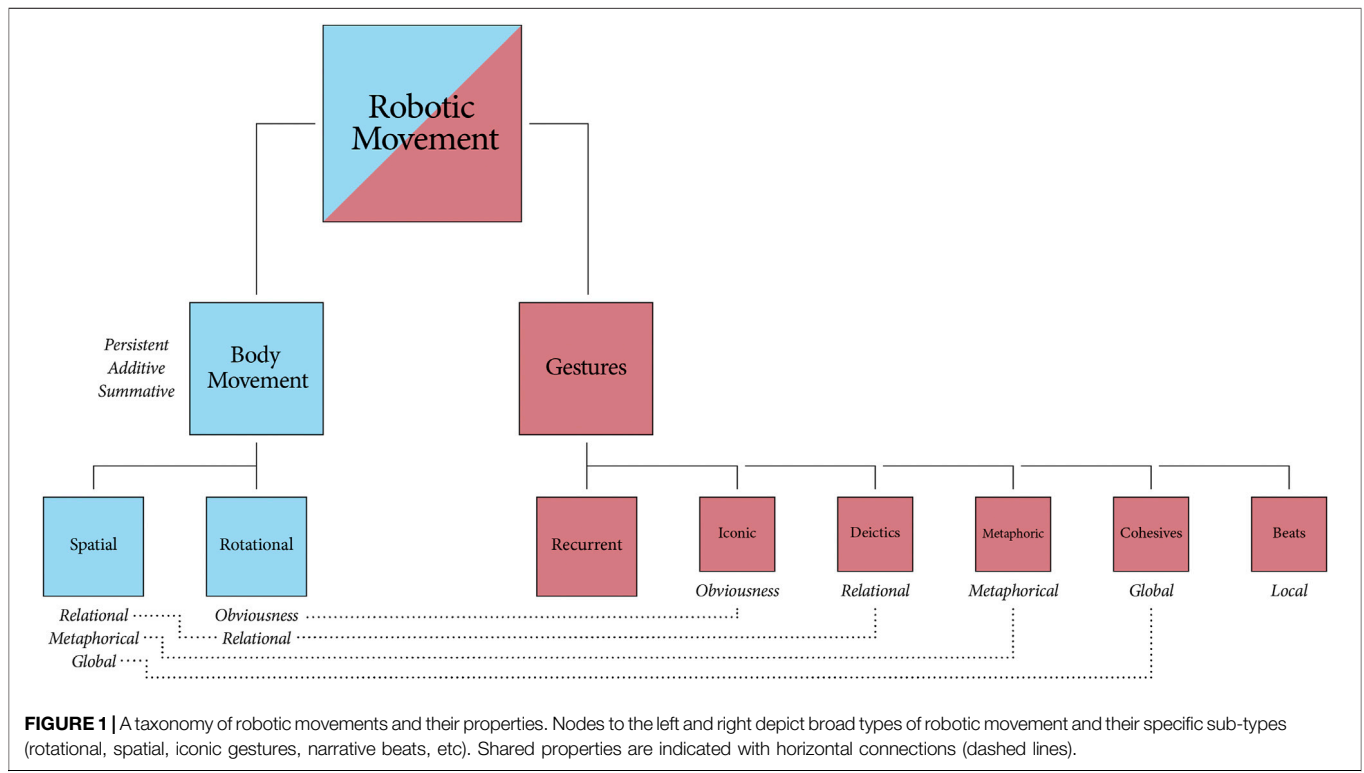

FIGURE 1 | A taxonomy of robotic movements and their properties. Nodes to the left and right depict broad types of robotic movement and their specific sub-types (rotational, spatial, iconic gestures, narrative beats, etc). Shared properties are indicated with horizontal connections (dashed lines).

## 2 BACKGROUND AND RELATED WORK

Although this framework is evaluated in a storytelling setting, it is applicable in different contexts of choreographed robotic interaction. Thus, we apply a new interpretation to existing data and argue that the framework has interdisciplinary relevance to other researchers in robotics.

Robots can make use of a variety of different modalities, each of which has been studied in different contexts: gaze (Mutlu et al., 2012; Andrist et al., 2014), facial expression (Reyes et al., 2019; Ritschel et al., 2019), voice (Niculescu et al., 2013), gesture (Pelachaud et al., 2010; Ham et al., 2011) and movement (Shamsuddin et al., 2011). The focus of our framework is on the movement of the robot, including both gestures and spatial movements (walking to and fro). Gestures have been extensively studied in linguistics and human-robot interaction, while spatial movement that concerns the whole body has been studied with social robots for comedy (Katevas et al., 2014), theater/improvisation (Bruce et al., 2000; Knight, 2011), and dance (LaViers and Egerstedt, 2012; Seo et al., 2013). Our consideration of related work thus takes an interdisciplinary look at various definitions and properties of gestures and holistic body movements, and derives a basis for characterizing the properties of spatial movements and gestures with reference to a series of empirical studies. We conclude by outlining the advantages of this framework for robotic performance.

### 2.1 Robotic Movement

We distinguish two classes of performative movement: local movement or gesture, typically with the arms, hands or head, and spatial relocation of the whole body. Gestures can arguably involve the whole body (as in e.g., air marshalling), while bodily locomotion can involve gestures while in motion. Yet the literature on gestures mostly confines gestures to the upper torso, arms and head (McNeill, 2008), while McNeill's widely used reference frame for gesture space depicts a sitting human with only the upper torso, arms and head in play (McNeill, 1992). Restriction to the upper body implies a locality of movement, while shifting the body in space has proximity effects that are global and relational. Locality and relativity are just two of multiple properties that reinforce a distinction between locomotive movements and gestures.

The taxonomic diagram in **Figure 1** defines the specific gestures and body movements we consider for our taxonomy. The top-most generalization *Robotic Movement* is split into *Body Movement* and *Gestures*, and each sub-type is linked to even more specific sub-types (vertically) and properties (horizontally). Although presented top-down, the taxonomy is a bottom-up approach that builds on a schematic basis for movement types to derive combinations and well-defined tools for roboticists and gesture researchers. Some related studies have looked at human motion in order to model movement dynamics (Bregler, 1997; Del Vecchio et al., 2003), while others have utilized computational models to simulate movement styles (Brand and Hertzmann, 2000; LaViers and Egerstedt, 2012).

#### 2.1.1 Gestures

In addition to considering the spatial trajectories of gestures, we must also look at their expressive role in communicating meaning. Empirical work by McNeill (1985), Bergen et al. (2003) and Hauk et al. (2004) has shown that gestures are an important instrument of human communication. It has also been argued that gestures are always embedded in a social, ideological and cultural context, and as such, they infuse our conversations with a contextual semantics (Bucholtz and Hall, 2016). Although some researchers have proposed a unified methodology for the semantic study of gestures (Mittelberg, 2007), there is as yet no clear consensus around a single framework. Studies that focus on the timely execution of gestures, such as those exploring gesture recognition (Kettebekov and Sharma, 2001; Sharma et al., 2008), follow Kendon's approach to the separation of gestures into *preparation*, *stroke* and *retraction* phases (Kendon, 1980). Here, we note that the necessary *retraction* after a gesture makes the gesture transient and ephemeral, so that the posture of the performer is the same before and after the gesture is performed. A broader classification which has been adopted in many studies is provided by McNeill (1992):

- **Iconic:** A gesture resembles what is denotes. Example: Shadow boxing when talking about a fight.
- **Deictics:** A pointing gesture may refer to another object. Example: Pointing at another actor on stage.
- **Metaphoric:** A figurative gesture should not be taken literally, yet it communicates a truth about the situation. Example: Showing a trajectory with the hand when talking about a trip.
- **Cohesives:** A cohesive gesture binds two temporally distant but related parts of a narrative. Example: Making the same hand movement whenever the same character appears.
- **Beats:** A gesture marks narrative time. Example: A rhythmic arm movement indicates time passing.

The class of *Iconic* gestures requires that users recognize the iconicity of a gesture when it is performed by a robot. A study by Bremner and Leonards (2016) shows that iconic gestures performed by a robot can be understood by humans almost as well as those performed by humans. Another study, conducted by Salem et al. (2011), suggests that human evaluation of a robot is more positive when it uses iconic, referential and spatial gestures in addition to speech. Regarding spatial and referential gestures, it has been argued that gestures are primarily used to augment non-visuospatial speech communication with visuospatial information (McNeill, 1992). In the five classes of gesture above, most can convey some visuospatial information, but *Deictic* gestures do so by definition. *Deictics* play a crucial role in human to human communication by supporting direct reference to visual and non-visual objects (Norris, 2011). It has also been shown that robotic deictic gestures can shift our attention in much the same way as human uses of these gestures (Brooks and Breazeal, 2006). The level of abstraction in *Metaphoric* gestures is generally higher than that of *Iconic* and *Deictic* gestures, and there is evidence to suggest that distinct integration processes apply to these different classes of gesture in the human brain (Straube et al., 2011). Metaphors exploit familiar source domains, so the same gestural movement can be metaphorical in one speech context and iconic in another. For example, the gesture "*raising one arm above the head with a horizontal, open hand*" is iconic when it accompanies the sentence "*The plane flew way above the clouds*", and metaphorical when it accompanies "*She is way out of your league*". A study by Huang and Mutlu (2013) investigated four of McNeill's gesture classes (all but *Cohesives*) as used by interacting humanoid robots in a narrative context. Those authors evaluate each gesture type on several fronts: information recall, perceived performance, affective evaluation, and narration behavior. In their study, *Deictics* are shown to improve information recall relative to other gestures, while *Beats* lead to improvements in effectiveness.

There are observable overlaps between the reference framework used within spoken language and the reference framework used with gestures (Cienki, 2013a). For example, if an event occurs to the left of a person, that person is more inclined to gesture to their left when retelling the event (McNeill, 1992). While this appears to hold for most Indo-European languages, there are some culturally dependencies. Speakers of the Mayan language Tzeltal use an absolute spatial reference framework for both speech and gesture, so if an event occurs to the west of a Tzeltal speaker, they are inclined to point west when they later tell of it (Levinson, 2003). Another example of cultural diversity is found in the Aymaran language. The Aymara people of the Bolivian Andes refer to future events by pointing *behind* rather than *ahead* of themselves (Núñez and Sweetser, 2006). However, some gestures appear relatively stable across cultures when there is a consistent, well-established link from form to meaning (Ladewig, 2014). These recurrent gestures often serve a performative role, and fulfill a pragmatic function when they work on the level of speech (Müller et al., 2013). We exclude this class of gestures from our movement framework, since we focus here on gestural meaning-making that is parallel to, and not so easily tangled up with, speech. Nonetheless, for the sake of completeness, the recurrent gestures are depicted atop the other gesture classes in **Figure 1**.

A variety of studies have looked at gestures in human-robot interaction. Ham et al. (2011) evaluated a storytelling robot with a set of 21 handcrafted gestures and 8 gazing behaviors. Csapo et al. (2012) presented a multi-modal Q&A-dialog system for which they implemented 6 discourse-level gestures, much as Häring et al. (2011) had earlier presented a multi-modal approach that included 6 specific upper-body postures. Those implementations use a small set of gestures, whereas others have made use of a reusable database with about 500 annotated gestures (Vilhjálmsson et al., 2007; Pelachaud et al., 2010). Those authors also describe a Behavioral Markup Language that allows virtual and physical presenters to use and combine these gestures. Despite sharing the goal of non-verbal communication with robots, most studies define gesture sets which are either specific to the task or to the robot. While the Behavioral Markup Language aims to overcome the latter, the iconicity and cultural-dependence of most gestures makes it is difficult to see how the implementation of task-specific gestures can be easily generalized.

### 2.1.2 Schematic Movement
We can also explore commonalities among gestures with regard to their embodied semantics. Cienki (2013a) argues that gestures ground the cognitive model of situated speakers in their physical environment. The schematic nature of certain movements across different gesture types has been related to image-schematic structures. These are recurring cognitive structures shaped by physical interaction with the environment (Lakoff, 2008; Johnson, 2013), and can be observed not just in verbal but in non-verbal communication (Cienki, 2013b; Mittelberg, 2018). Johnson provides the example "*Let out your anger*" (Johnson, 2013). Here, anger, a metaphorical "fluid" housed in the body, is said to be released from its container. As shown in (Wicke and Veale, 2018c), such schemas can be used to depict causal relations in embodied storytelling with robots. Not only does the theory of image schemas provide a *Conceptual Scaffolding* (Veale and Keane, 1992) for narrative processes, it also provides an algebraic basis for modeling complex processes and situations (Hedblom, 2020). In this way, simple schemas can be used as primitive building blocks of larger, more complex structures (Veale and Keane, 1992; Besold et al., 2017). For example, Singh et al. (2016) describe a playful co-creative agent that interacts with users by classifying and responding to actions in a 2D virtual environment. These authors train a Convolutional Neural Network on schematic movements so that it can classify inputs as, for instance, *Turn*, *Accelerate* or *Spin*. Moving from a two-dimensional plane to three-dimensional embodied space allows us to combine gestures with other body movements that extend beyond gesture space. Those extended movements can also tap into our stock of embodied schemas to support a metaphorical understanding of physical actions.

### 2.1.3 Body Movement

An advantage of the schematic approach is that a small set of robotic movements can produce a large number of useful combinations. For our current purposes we define just two types of bodily movements:

- **Spatial:** Movement along one axis
- **Rotational:** Rotation around one axis

Each type of movement fulfills a physical function: Spatial movement changes the position of the robot in space, while rotational movement changes the direction the robot is facing. It is known from the early studies of Heider and Simmel (1944) that even simple movements can lead an audience to project intentional behavior onto inanimate objects, to perceive emotion where there is only motion. A study by Nakanishi et al. (2008) shows that even minimal movement on one axis of a robot-mounted camera increases one's sense of social telepresence. Implementing rotation and directional movement in a museum robot, Kuzuoka et al. (2010) show that a robot's rotation can influence the position of a visitor, and that full body rotation is more effective than partial, upper-body rotation. Nakauchi and Simmons (2002) have investigated literal spatial movement in the context of queuing in line, and consider relative positioning in line as a parameter for achieving optimal, socially-accepted movement. Our focus here is on the metaphorical potential of bodily movement in a robotic context that must speak to human emotions. **Table 1** provides examples of how schematic constructions, implemented simply with robots, can convey intention and emotion. Of course, even the *metaphors we live by* (Lakoff and Johnson, 2008) can brook exceptions. For example, UP may generally signify good, and DOWN bad, but we want a fever to go down, and do not want costs to go up. This observation also applies to the schemas presented in **Table 1**. There are some situations where moving away increases emotional closeness, and moving closer decreases it, as when e.g., the former signifies awe and great respect, and the latter signifies contemptuous familiarity. As with all powerful schemas, we believe the benefits of generalization outweigh the occasional exceptions.

Following Falomir and Plaza (2020), who argue that primitive schemas like these can be a source of creative understanding in computational systems, we believe that simple schemas can be reused across creative applications of robotic movement, to connect the semantics of the task with the movement of the robots. Each movement may carry a unique semantics for different tasks, yet build on the same schematic basis. For example, the choreography of dancing robots can be synchronized using the same basic motions (back, forth, left, right). The dance can reflect abstract concerns through metaphorical motions, as when robots dance in a circle to reflect a repeating cycle of events. Likewise, in a storytelling context, actors can strengthen a perceived bond by moving closer together over the course of a story, or weaken and break that bond by gradually moving apart.

### 2.1.4 Limitations by Context

"Space" is a very general notion that can be understood in different ways in different performative contexts. For instance, our understanding of the movements of fellow pedestrians on the street is subtly different from our understanding of actors pacing about a designated stage. Even stages differ, and a proscenium arch can frame the action in a way that encourages a different kind of dramatic interpretation than a stage that is not so clearly divided from the viewing gallery. So our perception of how space is framed can influence our construal of meaning within that space (Fischer-Lichte and Riley, 1997). Just as the physical stage frames our conception of space, physical actors frame our notions of gesture and locomotion. We make a simplifying assumption in this work that our robots exhibit comparable degrees of freedom to an able-bodied human actor, but this need not be the case, Different robot platforms presuppose different kinds of movements, and support different degrees of physical verisimilitude (see **Section 6.3**). While we accept the limitations of our current platform, the anthropomorphic Nao, and choose our actions and schemas to suit these limitations, other robot platforms may afford fewer or greater opportunities for embodied meaning-making.

## 2.2 Exploring Meaning in Movement

Robots do more than stand in for the characters in a story. Their performances should convey meaning that augments that of the spoken dialogue and narration. When we speak of the semantic interpretation of movements and gestures, it is tempting to ground this interpretation in a componential analysis, and ask:

TABLE 1 | A listing of image schemas with their robotic realizations, with additional potential for metaphorical meaning. Each row contains a schema and its inverse.

| Schema | Movement | Metaphorical Meaning | Schema (Reverse) | Definition | Metaphorical Meaning |
|---|---|---|---|---|---|
| NEAR | The robot is moving near another robot or object | There is an interest or sympathy towards the robot/object | FAR | The robot is moving further away from the robot/object | There is a growing dis-interest or disliking to-wards the robot/object |
| FRONT | The robot is moving or turning in front of itself or another ro-bot/object | The robot is actively en-gaging with the other ro-bot/object | BACK | The robot is moving or turning to the back of itself or another robot/object | The robot is actively dis-engaging with the other robot/object |
| UP | The robot is moving upwards | The robot is displaying some super-iority over the other robot/object | DOWN | The robot is moving down or downwards | The robot is displaying some infer-iority over the other robot or object |

what are the components of gestures and other movements that convey specific aspects of meaning? In sign language, for instance, signs have a morphemic structure that can be dissected and analyzed (Padden, 2016). But gestures are not signs in any sign-language sense, and cannot usually be dissected into smaller meaningful parts. Indeed, signers can use gestures with sign language, just as speakers use them with spoken language (Goldin-Meadow and Brentari, 2017). Moreover, there is some neuro-psychological evidence that speech-accompanying gestures are not processed by language-processing mechanisms (Jouravlev et al., 2019). Our gestures give additional context to speech, while speech gives a larger context to our gestures. They add meaning to language (Kelly et al., 1999; Cocks et al., 2011) while not strictly constituting a language themselves. Some gestures indicate that a speaker is looking for a certain word [these are called *Butterworth* gestures by McNeill (1992)], and so serve a meta-communicative function. Likewise, a speaker can produce many kinds of unplanned movement while communicating with language, such as tugging the ear, scratching the head or waving the hands, and although an entirely natural part of embodied communication, we do not seek to replicate these meta-communicative gestures here. Rather, our focus is on gestures that communicate specific meanings, or that can be used to construct specific metaphors.

## 2.3 Creative Robots in Other Performative Contexts

Story-telling is just one performative context in which robotic actors use space and movement to convey meaning. For instance, robots have been used for improvisational comedy. *ImprovBot* (Rond et al., 2019) collaborates with human actors in ways that require it to spin around, move in circles, or move forward, backward, and sideways. Similarly, the robotic marimba player *Shimon* (Hoffman and Weinberg, 2011) recognizes the gestures of a human collaborator, and uses a schematic understanding of those gestures to make corresponding music-making decisions. A robot artist that creates photo montages and digital collages by interacting with a human user is discussed in Augello et al. (2016a). The robot takes its cue from a variety of information sources, one of which is the posture of its user. A design for a robot artist that interacts with a human user in a therapeutic setting is sketched in Cooney and Menezes (2018). Again, the aim is integrate a range of cues, both verbal and physical, from the human into the robot's physical actions. Just how well robots like these mesh with their collaborators, whether human or artificial, is the basis of the interactional "fluency" explored by Hoffman (2007).

When creative robots use motion to convey meaning, we expect them to aim for more than the "mere execution" of a literal script. Augello et al. (2016b) explicitly make the latter their goal, in the context of a robot that learns to dance in time to music. Another dancing robot system, that of Fabiano et al. (2017), chooses its actions to match the schematic drawings of dance movements shown to it by a collaborator.

Robot actors on a stage can be likened to human actors in a stage play, or to animated cartoon characters. In each case, however, the artifice succeeds to the extent that movements are considered natural. Laban movement analysis (LMA), which allows one to characterize the effort required for different bodily movements, in addition to modeling the body's shape and use of space, has been used by Bravo Sánchez et al. (2017) to support natural robot movements in short plays. Robots can enact artificial stories generated by an AI system, or they can interactively enact a human-crafted story. The *GENTORO* system of Sugimoto et al. (2009) does the latter, to encourage story-telling in children by combining robots and handheld projectors. A story-telling (or story-enacting) robot can be a physical presence, or a wholly virtual one, as in Catala et al. (2017). Nonetheless, results reported in Costa et al. (2018) show that embodied robots garner more attention and engagement that virtual ones.

As noted in (Augello et al., 2016b), a performer should do more than merely execute a script. Rather, it should interpret that which it sets out to perform, in whatever context – conversation, theatre, dance – it is designed to do so.

One of the first computational storytelling systems to consider context was *Novel Writer* Klein et al. (1973), a system for generating short tales of murder in a specific context (a weekend party). Simulation is used to determine the consequences of events as shaped by the chosen traits of the killer and his victim. Changing these traits can alter the simulation and produce different plot outcomes. Story-telling systems can obtain and set these plot-shaping traits in a variety of ways, both direct and indirect. A robot storyteller can, for instance, obtain personal traits from its users, by asking a series of personal questions that are shaped by its own notions of narrative (Wicke and Veale, 2018a). Those questions, and the answers that are provided, then shape the generated story, and provide a context for the audience to understand the actions of the performer. Basing a story on a user's own experiences is just one way of providing a clear interpretative basis for the performer's actions on stage. The more general approach provided here seeks to instead ground the performer's choices in a user-independent model of how characters are affected – and are seen to be affected – by the cumulative actions of the plot.

**TABLE 2 |** Examples of gestures: *Iconic*, *Deictic*, *Metaphoric*, *Cohesive* and *Beats* by a robot. Each gesture is ascribed a general property, along with its hardware requirements.

| Gesture Type (*Example*) | Depiction (*Example*) | Req. Hardware | Properties |
| --- | --- | --- | --- |
| Iconic (*Drive*) | 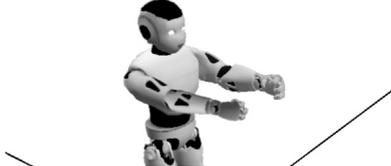 | Arms, Hands | Obvious |
| Deictic (*Point*) | 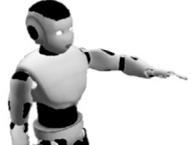 | Pointing limb | Relational |
| Metaphoric (*PUOH*) | 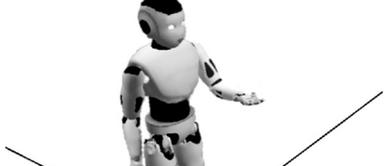 | Arms, Hands | Metaphorical |
| Cohesive |  | Limb | Global |
| Beats |  | Limb | Local |

# 3 PERFORMANCE FRAMEWORK FOR ROBOTIC ACTORS

## 3.1 Technical Description of Movements

The *Performance Framework* is applicable to a variety of embodied performance types that include robots, such as dancing, storytelling, joke telling and conversation. While those tasks impose unique requirements for hardware and software, the framework provides a unified conceptual perspective. The next sections present the framework, and explain its terminology and its syntax for describing movement. We start with a technical description of the specific movements that can be derived from the conceptual organization of movement types. As in **Section 2.1**, we address gestures first, followed by body movements.

### 3.1.1 Gestures

The properties of the five types of gestures described in (**Section 2.1.1**) are listed in **Table 2**, along with illustrative examples of each gesture. We illustrate each gesture in its most iconic form, with the exception of the *Beats* and *Cohesive* gestures, since these are always specific to the temporal context in which they appear. As noted earlier, the *Iconic* and *Metaphoric* gestures can use the same movements to convey different meanings in different contexts.

**Iconic:** An iconic gesture has an obvious meaning, since an icon can clearly substitute for what it is supposed to represent [see Peirce (1902) and Mittelberg (2019); the latter provides a thorough linguistic discussion of signs, icons and gestures]. These icons of physical actions are schematic by nature, insofar as they enact patterns of embodied experience (Mittelberg, 2019). We therefore attribute the property of *Obviousness* to the *Iconic* gestures, since they make meanings more explicit, and leave little room for alternate interpretations. **Table 2** presents an example of a robot steering an invisible vehicle to iconically depict the act of *driving*. Despite their obvious iconicity, many iconic gestures can be culturally-specific. As discussed in **Section 2.1.1**, gestures that are obvious to the speakers of one language may be confusing, misleading and far from obvious to members of a different cultural or linguistic grouping.

**Deictic:** Since pointing gestures refer to spatial/physical landmarks, we ascribe to *Deictic* gestures the property *Referential*. The technical implementation of such a gesture requires a limb, ideally an arm, that can point at the target reference. It is also beneficial if the pointing is further supported by the head or gaze direction of the robot (Clair et al., 2011). As with *Iconic* gestures, *Deictic* gestures also overlap with metaphorical gestures in different contexts (e.g., pointing ahead of oneself to signal a future event). **Table 2** shows the example of a robot pointing ahead with its arm.

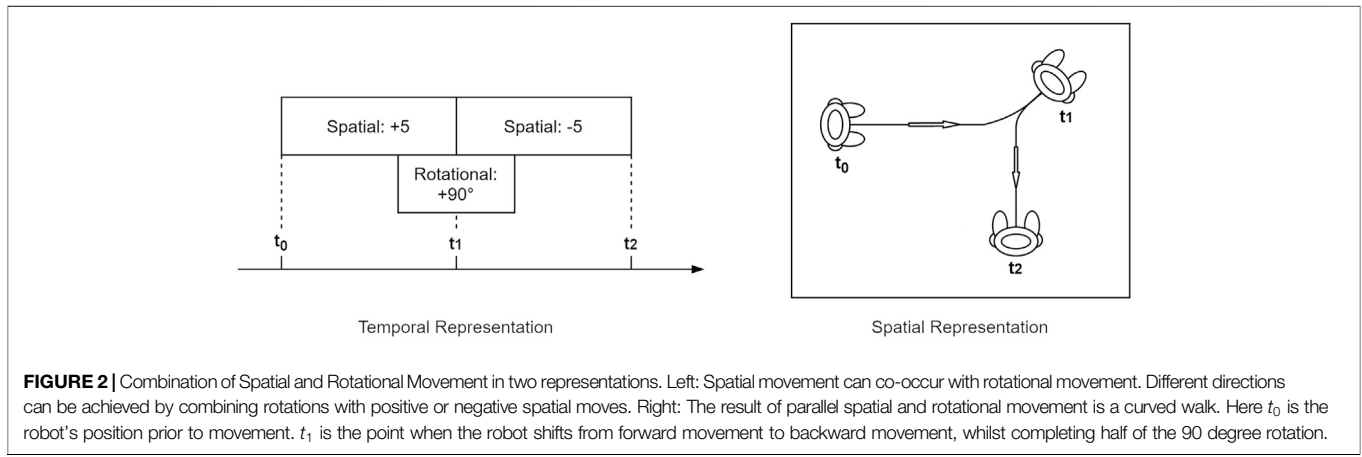

**FIGURE 2** | Combination of Spatial and Rotational Movement in two representations. Left: Spatial movement can co-occur with rotational movement. Different directions can be achieved by combining rotations with positive or negative spatial moves. Right: The result of parallel spatial and rotational movement is a curved walk. Here $t_0$ is the robot's position prior to movement. $t_1$ is the point when the robot shifts from forward movement to backward movement, whilst completing half of the 90 degree rotation.

**TABLE 3** | Depiction of the body movements *Spatial* and *Rotational* with their corresponding name, depiction with a physical robot, movement vector and properties in the respective columns.

| Movement | Depiction | Transformation Matrix | Properties |
|---|---|---|---|
| Spatial | 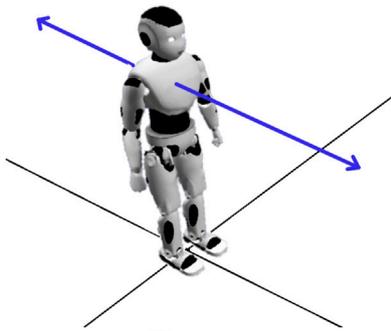 | $S_x(\omega) = \begin{bmatrix} 1 & 0 & 0 & 0 \\ 0 & 1 & 0 & 0 \\ 0 & 0 & 1 & 0 \\ \omega & 0 & 0 & 1 \end{bmatrix}$ | • Global<br>• Relational<br>• Summative<br>• Additive<br>• Persistent |
| Rotational | 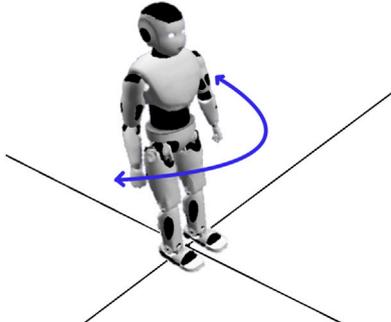 | $R_z(\alpha) = \begin{bmatrix} cos(\alpha) & -sin(\alpha) & 0 \\ sin(\alpha) & cos(\alpha) & 0 \\ 0 & 0 & 1 \end{bmatrix}$ | • Relational<br>• Obvious<br>• Summative<br>• Additive<br>• Persistent |

**Metaphoric:** *Metaphoric* gestures, labelled *Metaphorical* in **Table 2**, are the most challenging to implement since their intent must be discerned *via* a mapping from literal to non-literal meanings. Yet, as a consequence of this mapping, metaphorical gestures also open new possibilities for creativity within the system. An example of the creative use of metaphorical gestures is provided in **Section 4**.

**Cohesives:** Cohesive gestures are dependent on their context of use, and require careful timing. Whether a shaking of the fist, a circling of the finger or a turning of the wrist, such movements only make sense in a narrative if they are used coherently. Coherent usage aids discourse comprehension and allows audiences to construct a spatial story representation (Sekine and Kita, 2017). Moreover, *Cohesives* can strengthen our grasp of the whole narrative if they are used recurrently to reinforce persistent or overarching aspects of the plot. We therefore ascribe the attribute *Global* to these gestures.

**Beats:** *Beats* are just as context-dependent as *Cohesives*, but lack the latter's global influence, as they are relevant to one-off events only. Since the movement itself is less relevant than its timing and its context of use, no concrete example is offered in **Table 2**. In opposition to the *Cohesives*, we label these gestures as *Local*. Note that a gesture that is considered a *Beat* in one task domain, such as story-telling, might serve a global role in the synchronization of movement in another, such as dance. In that case, the gesture would be labelled *Global* in the latter context.

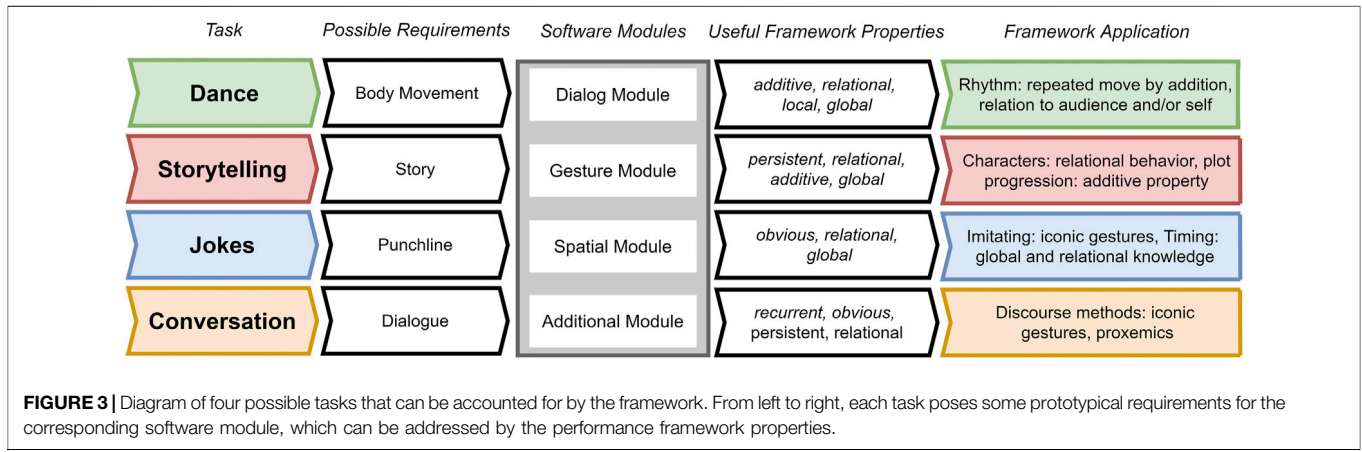

**FIGURE 3** | Diagram of four possible tasks that can be accounted for by the framework. From left to right, each task poses some prototypical requirements for the corresponding software module, which can be addressed by the performance framework properties.

Since the gesture types illustrated in **Table 2** do not constitute an exhaustive list, some additional properties may need to be included in the future. For example, some gestures are performed with two hands and so can exhibit relational properties in the way that each hand, representing a character, relates to the other (Sowa and Wachsmuth, 2009). We can also consider the naturalness of the gesture, as this is an important property for HRI and a common basis for assessing any computational model that uses gestures (Salem et al., 2012; Huang and Mutlu, 2013). However, we might also view naturalness – as we do here – as an emergent property of the implementation, rather than as a constitutive property of the performance framework itself.

### 3.1.2 Body Movement

**Spatial Movement** describes a simple trajectory of an agent along one axis. This core movement requires the agent to possess a means of locomotion, such as wheels or legs. In its basic form, spatial movement in one direction is a transformation of the positional coordinates in one variable:

$$\vec{x} = (\omega, 0, 0) \qquad \text{with } \omega \in \mathbb{R} \qquad (1)$$

The corresponding translation matrix is given in **Table 3**. This movement is compatible with all other movement types. Combined with rotational movement, it covers all directions on the 2D plane. When the mode of locomotion allows for it, the vector can be positive or negative. This kind of movement can exhibit the following properties:

- **Global:** The moving body affects the relative proximity, shared references and spatial configuration of all agents in a performance, and so has implications for the performance of the narrative as a whole.
- **Relational:** The movement has implications for other agents on the stage since an absolute change in position for one actor also changes its position relative to others.
- **Summative:** The movement of an actor into its resulting position summarizes, in some general sense, the history of past actions up to this point.
- **Additive:** A movement compounds a previous action to achieve a perceptible cumulative effect.

**TABLE 4** | Possible combinations of movement types. *comb.* (green) are **combinable** movements, *restr.* (yellow) are combinations that are only possible to some **restricted** extent and *excl.* (red) are movements that are mutually **exclusive**.

| Movement Type | Spatial | Rotational | Iconic | Deictic | Metaphoric | Cohesive | Beats |
|---|---|---|---|---|---|---|---|
| **Spatial** | comb. | comb. | restr. | restr. | restr. | comb. | comb. |
| **Rotational** | comb. | comb. | restr. | restr. | restr. | comb. | comb. |
| **Iconic** | restr. | restr. | comb. | comb. | excl. | comb. | excl. |
| **Deictic** | restr. | restr. | comb. | comb. | comb. | comb. | excl. |
| **Metaphoric** | restr. | restr. | excl. | comb. | comb. | comb. | excl. |
| **Cohesive** | comb. | comb. | comb. | comb. | comb. | comb. | excl. |
| **Beats** | comb. | comb. | excl. | excl. | excl. | excl. | comb. |

- **Persistent:** A movement has a lasting effect on an actor or its the physical relationship to others.

The property *Obviousness* is not attributed here, since actors (robotic or otherwise) can move in space without necessarily conveying meaning. Some movements help a speaker to communicate while being uncommunicative in themselves, as when an actor steps back to maintain balance, or moves their hands in time to their words as they speak. In contrast, it is hard to perceive a rotational movement as unintentional, since rotation carries such an obvious, iconic meaning. Thus, while the property of *Obviousness* is not wholly context-free, it is sufficiently robust across contexts to earn its keep in a performative HRI system.

**Rotational Movement** This movement requires an actor to possess some form of rotational joint, so that it can rotate around one axis. While a humanoid robot can simply turn its head or torso, this kind of rotational movement requires full body rotation. In some cases, rotation is only possible in combination with spatial movement. For example, some bipedal robots cannot rotate on the spot, and need to walk in a curve to achieve full rotation. The rotation around one axis is given by the transformation matrix in **Table 3** as $R_z(\alpha)$ (with α as degree of rotation). This kind of movement can exhibit the following properties:

- **Obviousness:** When the movement achieves the iconic action of turning away from, or turning toward someone else, this iconicity deserves the label *Obvious*.
- **Relational:** The rotation has implications for other agents on the stage since an absolute change in orientation for one actor changes its relative orientation to other agents.
- **Summative, Additive and Persistent:** These properties hold the same meanings for rotational movements as they do for spatial movements.

## 3.2 Combinations of Robotic Movement

Defining the basic types of movement and their properties provides a foundational set of movements that can be implemented for different kinds of robots. Basic movements can be considered primitive actions in a performance system, whose possibility space is the space of their possible combinations. Gestures can be combined with whole body movements (spatial and rotational) to produce complex behaviours. The individual movements themselves are not creative – many are iconic, and highly familiar – but the mapping from narrative to physical action does allow for metaphor and for other creative choices (Boden, 2004). The example combination provided in **Figure 2** shows a forward movement followed by a backward movement, paralleled by a rotational movement during the transition. The resulting performance (see **Figure 2** *Spatial Representation*) is the sum of its parts, and fosters audience interpretation of the performer's behaviour. This is where the properties *Summative*, *Additive* and *Persistant* come into play.

An embodied performance can draw on all available movements and all possible combinations of such. **Table 4** presents a combination matrix showing possible combinations, mutually exclusive movement types, and restricted combinations. The group that is least conducive to interaction with others is the *Beats*. Due to their *local* property, these are grounded in a specific narrative moment, which does not permit metaphorical, iconic or deictic displays. This momentary status also strongly prohibits combinations with *Cohesives*. In short, only *Beats* can combine with *Beats*. Nonetheless, *Beats* can be performed during spatial or rotational movements, as this does not change their function. In fact, spatial and rotational movements can be combined with all other movement types, as well as with themselves. However, a spatial or rotational movement during an iconic or metaphoric gesture can cause positional changes that affect the gesture, while deictic gestures are also sensitive to any referential changes of position. For example, pointing while walking is a much more restrictive task than either alone, since the target of the reference might move behind the performer.

By definition, iconic and metaphoric gestures exclude each other. As with a change of context, a gesture's obviousness can be exchanged for a metaphorical interpretation, but a combination of iconic and metaphoric gestures must be sequential, not parallel. Likewise, the *Cohesives* can be combined with any other movement type except for the *Beats*, since these groups have opposing *global* and *local* properties.

## 3.3 Performance Framework at Work

**Figure 3** depicts four example tasks, the requirements of each, and the applicability of the framework to each instance of the task. The framework is designed to meet the demands of these different tasks. When the *Software Modules* for a task depend on the choice of performing agents (e.g., embodied/non-embodied, single/multiple), the properties needed to support an appropriate conceptual response are given.

Different performances types can place varying emphases on the meanings of any given movement. Dancing is an expressive act which aims to convey themes and emotions through the use of the entire body. While **Figure 3** lists only *Body Movement* as a necessary requirement for dance, dancing can have other requirements in context, e.g. single or multiple bodies which can - but do not need to - move synchronously. Dance types can range from the highly coordinated to the highly improvised and relatively uncoordinated. As shown in **Figure 3**, rhythm can, for example, be achieved with a repetition of movements. However, while rhythm and synchrony are listed as prototypical requirements of a dance task, these are neither necessary nor sufficient for dance, and this point applies more generally to all performance types, from dance to storytelling to joke-telling and casual conversations. In robotic dance, complex relational movements and motion dynamics are at play, which may or may not exhibit synchrony and rhythm (LaViers and Egerstedt, 2012; LaViers et al., 2014; Thörn et al., 2020). Within the *Performance Framework*, additive, relational, local and global properties can be identified, and, in cases where it is required, rhythm can be achieved by adding repeated movements, just as synchronized movement can be realized in terms of global and relational additions. Ultimately, *Body Movements* is a flexible requirement which should always acknowledge the diversity of bodily capabilities across humans and across robots, making it all the more important that each possible requirement is appropriately integrated on the software level.

For storytelling, Wicke and Veale (2020b) have shown that movement, gesture and relative positioning play an important role in enacting a story well. These requirements can each be met using movements with persistent, relational, global and additive/summative properties. When actors undergo changes in their physical spaces that mirror the changes undergone by characters in a narrative space, metaphorical schematic movements and gestures can depict plot progression and character interrelationships.

Certain performance types, such as stand-up comedy, place a greater emphasis on timing than others (Vilk and Fitter, 2020). When timing is key, spatial movements may be subtle and minimal (Weber et al., 2018), making the properties *obvious*, *relational* and *global* all the more important. For example, the timing needed to land a punch line requires global and relational knowledge of the performance as a whole, while the use of iconic gestures throughout can increase the effectiveness of the performance.

Lastly, conversational agents make use of various discourse strategies that can be enhanced by the use of iconic and deictic gestures. The latter are especially useful in maintaining shared attention and awareness, by mirroring movement in a topic space with movement in physical space (Jokinen and Wilcock, 2014).

# 4 AN INTERPRETATION FRAMEWORK FOR STORYTELLING ROBOTS

By meaningfully connecting plot actions to movements, the *Performance Framework* allows a performer to pick its movements to suit the action ($x$) at hand. More formally, $C(x) \mapsto E(x)$ denotes the mapping of an action $x$ as it is represented in the conceptual domain $C$ of stories to its expressive realization in the embodied, physical world $E$. For example, the *insult* action can be expressed with an iconic gesture in which an actor "flips the finger" to another actor. That other actor may show that they feel disrespected by moving their head slightly backwards. In this case, $C(insult) \mapsto E(insult)$ because the actors physically express the *insult* action that the plot calls for. However, each actor should take into account the current state of the story, and their residual feelings that carry over from earlier actions. If we denote this state of the story as $S$, then the performers consider the mapping $S(C(insult)) \mapsto S(E(insult))$. In a story space with $x_N$ possible actions, the general form of this mapping is $S(C(x_N)) \mapsto S(E(x_N))$.

Skilled actors are able to interpret an action within the context of the unfolding story, so we also need a complementary *Interpretation Framework* to allow performers to interpret each action in context. Suppose character A has supported B in some way, or confided in B, or defended B, and B then responds by insulting A. Viewed in isolation, the insult should make A feel disrespected, and even a little attacked, so it would be appropriate to embody this event as $C(insult) \mapsto E(insult)$. However, given the earlier events which make this insult all the more shocking, it would be even more appropriate, from A's perspective, to enact $C(insult) \mapsto E(attack)$, since *attack* carries more shock value than *insult*. Each performer brings a different interpretation to bear on the same plot action. So while B interprets the *insult* action directly, A interprets it as *attack* action. The result is a performative blend of the two enactments. B enacts its agent role in the *insult* while A enacts its patient role in the *attack*. That is, while B enacts the event *via* the mapping $S(C(insult)) \mapsto S(E(insult))$, A uses the mapping $S(C(insult)) \mapsto S(E(attack))$. The more general form of A's interpretation is $C(x) \mapsto E(\bar{x})$. It is the task of the *Interpretation Framework* to provide the mapping mechanisms for interpretations such as these.

## 4.1 The Representation of Gestures Within the Framework

Wicke and Veale (2018b) define one-to-many mappings from plot actions to gestures and movements, from which performers can choose an appropriate but context-free enactment at random. The purpose of the performance framework is to transform this choice from a purely disjunctive one to a choice based on interpretation in context. To this end, each gesture must be understood by the system as more than a black box motor script. So in addition to duration information, a schematic classification as given in **Table 1**, and the properties given in **Table 2**, we must give the framework an emotional basis for making real choices.

Our database of gesture representations is available from a public repository[1]. More than 100 movements are currently stored and labelled for use in embodied storytelling. Each is assigned a unique name that describes the movement briefly. This name aims to be as telling as possible in just a few words, while a longer free-form description is as explicit and detailed as possible. For example, the movement named "strike down" has the description "right arm squared angle lifted above shoulder, quickly striking down with hand open." This movement, which takes approximately $2\frac{1}{2}$ seconds to execute, is labelled as a schematic *down* movement. This motion is not associated with rotational or spatial movement, and its possible uses as an iconic or metaphorical gesture depend on the narrative context in which it is performed.

As noted in (Wicke and Veale, 2018b), the existing disjunctive mapping from plot actions to physical actions is further labelled with an appropriateness label, since some gestures are more obviously suited to their associated plot actions than others. For example, the action *disagree_with* can map to either of the gestures "shaking the head" or "shaking the head, raising both arms and turning away." In this case, both gestures are equally appropriate for the action. For another action, however, such as *contradict* or *break_with*, the latter is more appropriate than the former. Three distinct appropriateness levels – *high*, *medium* and *low* – are used to qualify the mappings of actions to gestures. This suggests a very practical motivation for metaphor within the system: the mapping $C(x) \mapsto E(\bar{x})$ is preferable to $C(x) \mapsto E(x)$ when $E(\bar{x})$ offers a more appropriate enactment for $x$ than $E(x)$. In general, metaphor will be motivated by a mix of concerns, from the practical (does this action have a vivid enactment that really suits it?) to the expressive (does this action adequately capture the feelings of the moment?). Notice that in each case, however, metaphor hinges on questions of expressive adequacy, and the question of whether the systems knows of a better way to communicate what it wants to say.

---

[1] https://osf.io/e5bn2/?view_only=2e30ee7e715342d59c371b5d30c014e0

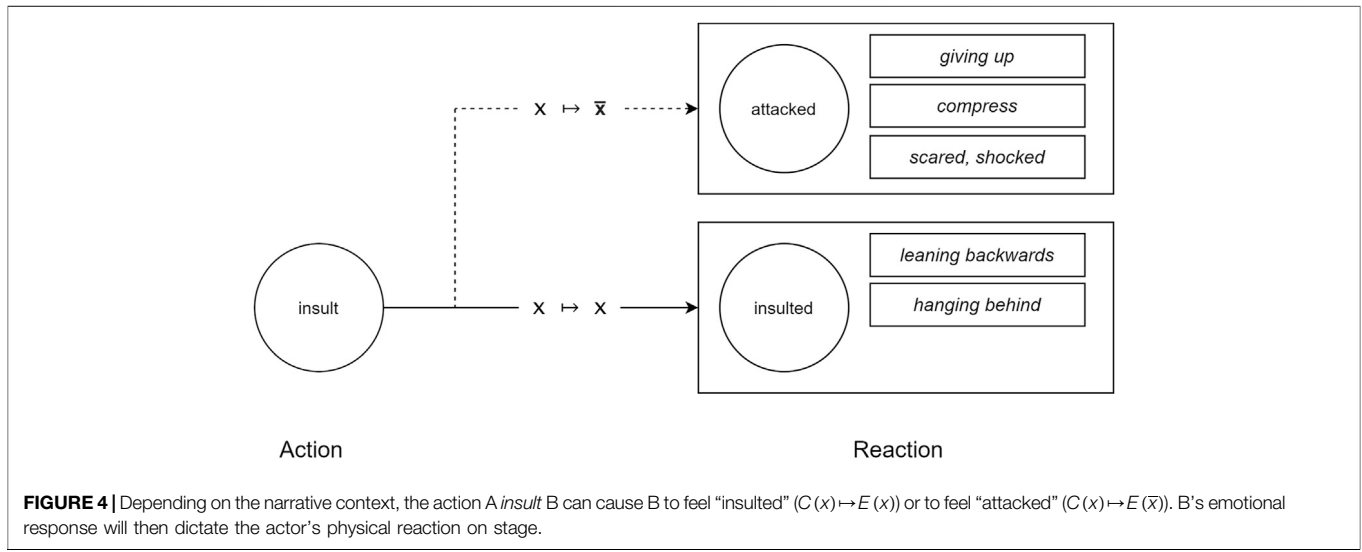

**FIGURE 4** | Depending on the narrative context, the action A *insult* B can cause B to feel "insulted" ($C(x) \mapsto E(x)$) or to feel "attacked" ($C(x) \mapsto E(\bar{x})$). B's emotional response will then dictate the actor's physical reaction on stage.

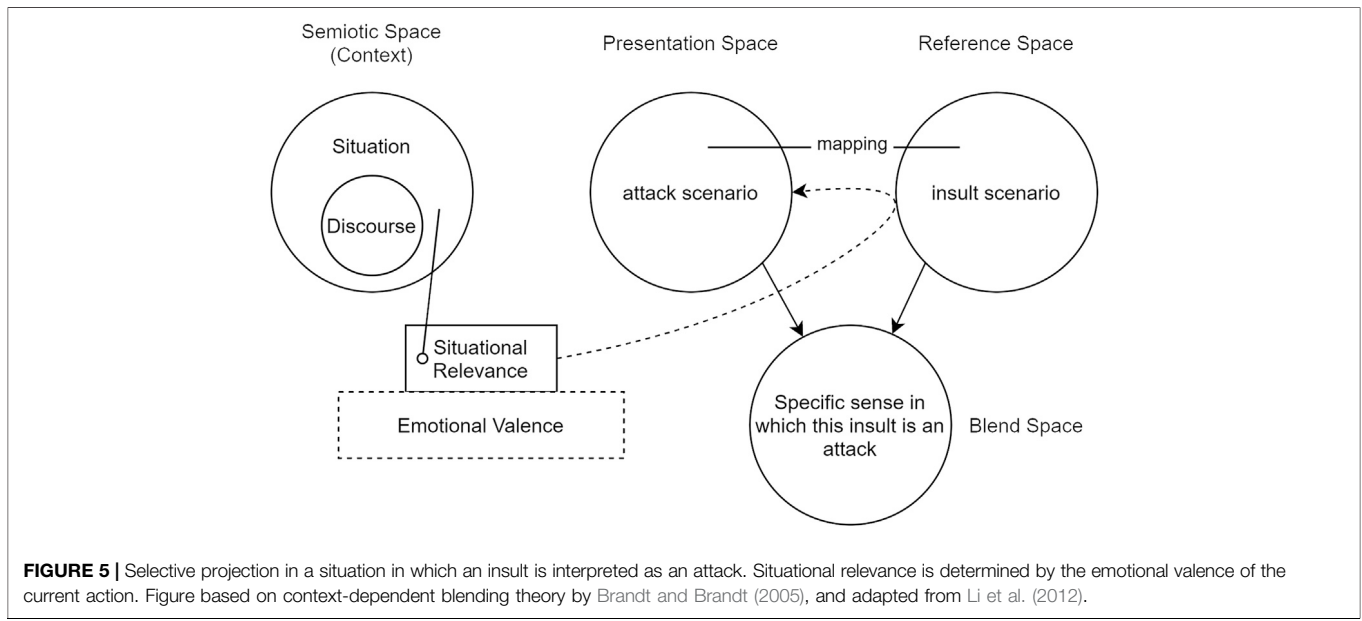

**FIGURE 5** | Selective projection in a situation in which an insult is interpreted as an attack. Situational relevance is determined by the emotional valence of the current action. Figure based on context-dependent blending theory by Brandt and Brandt (2005), and adapted from Li et al. (2012).

## 4.2 Selective Projection for Creative Interpretation

An embodied performance of a story is a careful presentation of story elements – plot, character, emotion – in a physical space. As such, performers project elements from the story space, a space of words and concepts, into the presentation space, a space of gestures and movements and spoken dialogue. The performers are themselves, with their own physical affordances and limitations, *and* the characters they play. In the terminology of Fauconnier and Turner (1998), Fauconnier and Turner (2008), the performance is a *conceptual blend*.

Turner and Fauconnier's *Conceptual Blending Theory* has previously been used to model stories in a computational setting (Li et al., 2012). The basic theory has been extended by (Brandt and Brandt, 2005) to incorporate additional spaces that are especially relevant here, such as a reference space (for the underlying story), a context space (specifying situations within the story, and discourse elements relating to those situations), and a presentation space in which story elements are packaged and prepared for a performance.

Consider again the example story in which character B insults A after A has shown favor to B, perhaps by praising, aiding or defending B. In the reference space this plot action is literally captured as *B insult A*. As mediated by the context space, however, which brings both situational relevance and a discourse history to the interpretation of events, B views this insult as an attack, and so the action is instead represented in the presentation space as *B attack A*. Since the performers take their stage directions from the presentation space, A will move, gesture and speak as though the victim of an actual attack. So, when B performs a "giving the finger" gesture to A, A will do more than lean back in disappointment – the standard response to an insult – it will step away with its arms extended in a defensive posture. This construal of events by A and B, the first of three scenarios unpacked below, is illustrated in **Figure 4**.

**Scenario 1**: An insult delivered in some contexts can surprise more, and wound more, than in others. The standard response, which entails a literal mapping from the reference to the presentation space, is $S(C(insult)) \mapsto S(E(insult))$. However, in a story state $S$ that makes the insult seem all the more severe, $S(E(insult))$ may equate to $E(attack)$, to produce the non-standard mapping $S(C(insult)) \mapsto E(attack)$. In that case, it is not the embodied response to an insult, $E(insult)$ that is enacted by the insult's target (leaning back, with head down) but $E(attack)$ (stepping back, arms outstretched defensively).

**Scenario 2**: A performer whose character, A, praises the work of another, B, might enact a show of "praise" with a clap of the hands or a nod of the head. This is the standard response in a story context where praise is literally interpreted as praise, that is, $S(C(praise)) \mapsto S(E(praise))$. However, if the context indicates that A has strong grounds to respect and feel inspired by B – perhaps B rescued A in the previous action – then $S(E(praise))$ may be interpreted in this light to produce a stronger reaction than praise. As such, $S(E(praise))$ might equate to $E(worship)$ and the performer playing A will bow accordingly.

**Scenario 3**: A succession of actions that reinforce the same emotional response in a character can lead to a character feeling and expressing that emotion to a higher degree, shifting its embodied response from the standard interpretation to a heightened, metaphorical level. Suppose the story concerns character A treating character B as a lowly minion. A overworks and underpays B, taking advantage of B at every turn. If A should now scold B, B may interpret $S(E(scold))$ as $E(whip)$, or interpret $S(E(command))$ as $E(enslave)$, and finally interpret $S(E(fire))$ as $E(release)$. Interpretative performance allows for a shadow narrative to play out in physical actions as the literal narrative is rendered in speech.

In each scenario, the situated actor uses contextual information to interpret the current plot action, in light of previous actions, and chooses to accept the scripted action $(x \mapsto x)$ or to take a metaphorical perspective $(x \mapsto \overline{x})$ instead. An alternate construal, such as construing an insult as an attack, or an act of praise as an act of worship, changes the physical enactment of the action in the script. Notice that when an alternate enactment is chosen, the dialogue associated with the scripted action is still used. The combination of one action's gestures with another action's dialogue adds further variety to the blend, while also helping to foster understanding by the audience. Gestures are dramatic on a physical level, but dialogue carries a more explicit semiotic content. Even when the performers choose to be metaphorical, the performance remains grounded in some literal aspects of the script. This grounding is rooted in the assumption that audiences are capable of fully comprehending the narration and dialogue of the script. When this is not the case, gestures and other non-verbal cues become an even more important channel of communication.

A blending interpretation of *Scenario 1* is illustrated in **Figure 5**, further adapting the treatment of Brandt and Brandt (2005) that is offered in Li et al. (2012). Notably, situational relevance is informed by the *Emotional Valence* of the situation, the calculation of which we consider next.

## 4.3 Emotional Valence in Story Progression

Veale et al. (2019) have shown how the actions, characters and structural dynamics of a story can influence the performers' reactions so as to elicit a comedic effect in a performance by robots. In that approach, the logical structure of the narrative – in particular, whether successive actions are linked by "but" or "then" or "so" – provides a reasonable substitute for an emotional interpretation of the action, so that performers know when to act surprised, or can infer when an audience might be getting bored (e.g., because the plot lacks "but" twists) or confused (e.g., because it has too many "but" twists). To go deeper, we must augment this structural perspective with an emotional perspective, so that performers can grasp why certain actions are linked by a "but" and not a "so". We begin by situating each role (A and B) of every possible action in a plot (*Scéalextric* defines more than 800 different actions) on the following four scales:

$$\text{disappointed} \leftarrow A \rightarrow \text{inspired} \quad \text{disappointed} \leftarrow B \rightarrow \text{inspired}$$
$$\text{repelled} \leftarrow A \rightarrow \text{attracted} \quad \text{repelled} \leftarrow B \rightarrow \text{attracted}$$
$$\text{attacked} \leftarrow A \rightarrow \text{supported} \quad \text{attacked} \leftarrow B \rightarrow \text{supported}$$
$$\text{disrespected} \leftarrow A \rightarrow \text{respected} \quad \text{disrespected} \leftarrow B \rightarrow \text{respected}$$

These emotions are chosen to suit the action inventory of a story-telling system like *Scéalextric*. Other emotional scales may be added as needed to suit other tasks, such as dance (Camurri et al., 2003). The story-telling system draws from a knowledge base of over 800 actions, which can be causally connected to create stories that exploit tropes and other common narrative structures. Each story revolves around two central characters and a retinue of secondary figures (partners, spouses, friends, etc.). The most common themes elicit feelings of trust, respect, admiration and cooperation about and between those characters.

For example, the *insult* action associates a strong sense of being disrespected with the B role. When A insults B, we expect B to feel very disrespected (or negatively respected). Similarly, the *worship* action associates a strong sense of being inspired with the A role, and a strong sense of being respected with the B role. Conversely, the *surrender to* action associates a negative sense of being attacked (and so a positive sense of being supported) with the B role, because A is no longer an active threat to B. Viewed individually, each emotional setting can be compared to that of the previous action, to determine how much change has been wrought by the current plot turn. It is this change that explains why certain transitions warrant a "but" and others warrant a "so" or a "then". It can also motivate why an insult can come as a surprise to a character, and feel more like an attack.

The four emotional scales can also be viewed in the aggregate, to determine an overall valence for the current action from a given role's perspective, or to determine an overall shift in valence from one action to the next. We calculate the valence of a role in an action $\alpha_i$ as the total valence across all emotional scales for that role in that action. See **Eqs 2**, **3** for the valence of the A and B roles in $\alpha_i$. A positive valence for a role indicates that a character in that role should experience a positive feeling when the action is performed; conversely, a negative valence suggests a negative feeling for the action.

$$valence_A(\alpha_i) \leftarrow inspiration_A(\alpha_i) + attraction_A(\alpha_i) + support_A(\alpha_i) + respect_A(\alpha_i) \quad (2)$$

$$valence_B(\alpha_i) \leftarrow inspiration_B(\alpha_i) + attraction_B(\alpha_i) + support_B(\alpha_i) + respect_B(\alpha_i) \quad (3)$$

A character is a persistent entity in a narrative, one that moves through the plot from one action to the next. The current valence of a character is a function of the valence of the role it plays in the current action, and of the valence of its roles in previous actions, with the current action making the greatest contribution. Previous actions have an exponentially decaying effect based on their recency. If $0 < \beta < 1$ specifies the weight given to the current action, the contextual valence of the characters filling the A and B roles is given by **Eqs 4**, **5** respectively. We assume a fixed decay rate, while acknowledging that certain events might have a stronger and more lasting impact on perceived valence than others. It remains to be seen in future work whether this simple one-size-fits-all approach needs to be replaced with a more variable, local solution. For now, we continue to view this simplicity as a virtue.

$$context_A(\alpha_i) \leftarrow \beta.valence_A(\alpha_i) + (1-\beta)context_A(\alpha_{i-1}) \quad (4)$$

$$context_B(\alpha_i) \leftarrow \beta.valence_B(\alpha_i) + (1-\beta)context_B(\alpha_{i-1}) \quad (5)$$

Calculating aggregate valence levels for the characters in a story allows the interpretation framework to track their changing emotions to each other over time, at least on a gross level. Although it is highly reductive, this gross level allows performers to distil complex emotions into simple but expressive physical actions. Because they are calculated as a function of the valence of current and past actions, these levels are both summative and persistent, and thus well-suited to making decisions regarding summative and persistent physical actions in a performance. If a significant increase in positive valence for a character A is interpreted as a result of actions involving character B, then performer A can move a step closer to performer B in physical space. Conversely, a significant decrease can cause A to move a step away from B. This increase or decrease for A is given by **Eq. 6**. The same spatial/emotional calculus applies to B's perspective, as given in **Eq. 7**. In each case, a significant increase or decrease is determined to be a positive or negative change that exceeds a fixed threshold τ. In this way, the relative spatial movements of performers on stage are not explicitly indicated by the script, or directly associated with the actions in the plot, but determined by each performer's evolving interpretation of the narrative context.

$$\Delta_A(\alpha_i) \leftarrow context_A(\alpha_i) - context_A(\alpha_{i-1}) \quad (6)$$

$$\Delta_B(\alpha_i) \leftarrow context_B(\alpha_i) - context_B(\alpha_{i-1}) \quad (7)$$

The emotional valence of an action for a character, much like a character's "inertial" contextual valence, is derived from four emotionally charged scales that have been chosen to suit our system's inventory of 800 plot actions. New parallel scales can be added, or existing ones removed or replaced, if this inventory were to change significantly. Currently, one obvious omission is an arousal scale (Kensinger and Schacter, 2006), to show the degree to which an action either calms or arouses a particular role. Arousal is not a charged dimension – for one can be as aroused by hate as by love – and so it does not contribute to our calculations of valence. Nonetheless, an arousal dimension is useful for indicating the scale of an actor's response. A high-arousal action may demand a bigger and more dramatic physical response that a calming, low-arousal event. For that reason, it makes sense to add a new scale as follows:

$$calmed \leftarrow A \rightarrow aroused \quad calmed \leftarrow B \rightarrow aroused$$

A state of high-arousal can be conveyed with a sweeping, high-energy gesture. while a calm state might be conveyed with a slow movement or a slight gesture. Of course, the robot platform may not support the distinction between high- and low-energy motions. The extent to which it does, or does not, indicates the extent to which an arousal dimension is worthwhile in a story-telling context. Still, we may find that arousal is wrapped up with the question of contextual valence and how quickly the influence of context should decay. If arousal can be shown to influence the rate of decay, it would be a valuable addition to the framework whatever robot platform is used. It thus remains a topic of ongoing research in this project.

## 5 EVALUATION

When a performer's spatial movements and gestures are chosen on the basis of its interpretation of the plot, we deem those physical actions to be *coherent*. Conversely, when those movements and gestures at chosen at random, to create the mere appearance of embodied performance, we deem those actions *incoherent*. Clearly, the value of interpretation lies in the audience's ability to recognize coherent uses of movement and gesture. More importantly, it lies in the increased appreciation that an audience will feel for coherent vs. incoherent performances. This is what we set out to evaluate here, by asking: how much do audiences appreciate spatial and gestural embodiment, and how much do they appreciate the interpretation that goes into the embodied choices that are made by our robotic performers when enacting a story?

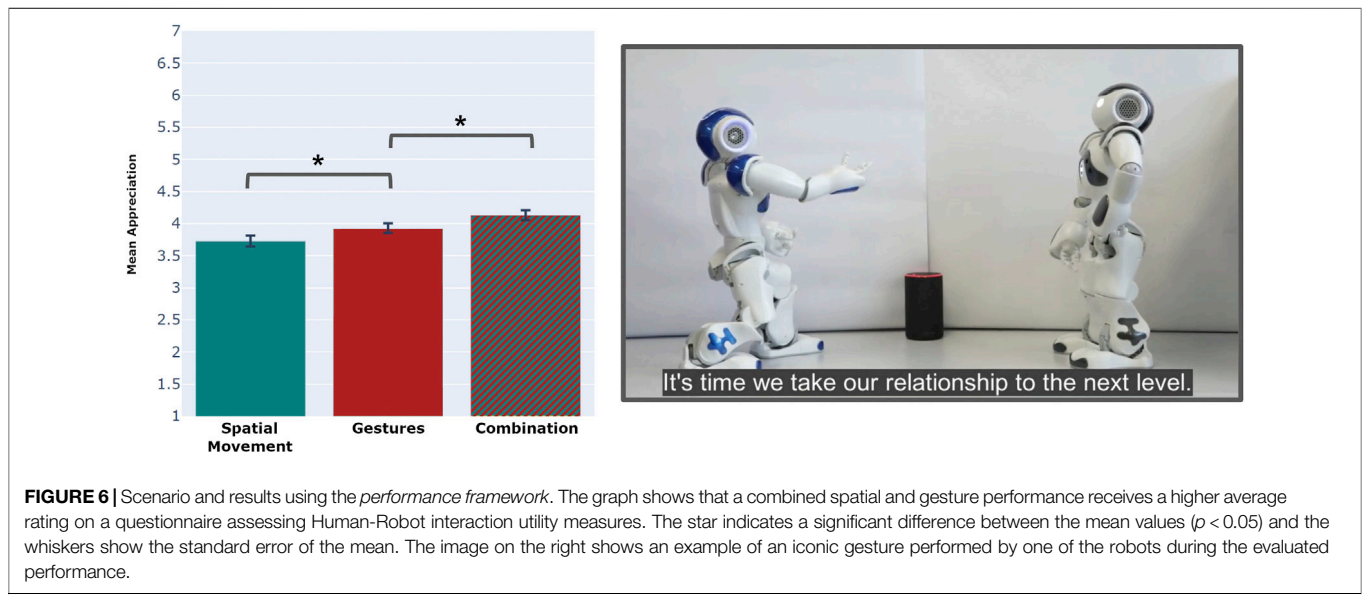

**FIGURE 6** | Scenario and results using the *performance framework*. The graph shows that a combined spatial and gesture performance receives a higher average rating on a questionnaire assessing Human-Robot interaction utility measures. The star indicates a significant difference between the mean values ($p < 0.05$) and the whiskers show the standard error of the mean. The image on the right shows an example of an iconic gesture performed by one of the robots during the evaluated performance.

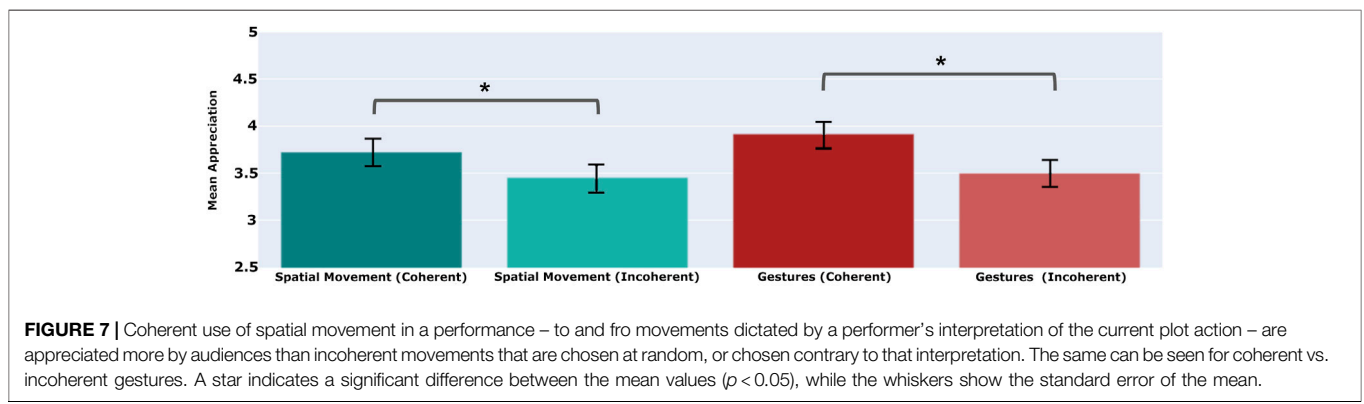

**FIGURE 7** | Coherent use of spatial movement in a performance – to and fro movements dictated by a performer's interpretation of the current plot action – are appreciated more by audiences than incoherent movements that are chosen at random, or chosen contrary to that interpretation. The same can be seen for coherent vs. incoherent gestures. A star indicates a significant difference between the mean values ($p < 0.05$), while the whiskers show the standard error of the mean.

## 5.1 Space and Gesture: Together and Apart

These stories are generated using *Scéalextric*, and performed by two Softbank NAO robots working within a related framework, named *Scéalability*, that choreographs their actions (Wicke and Veale, 2020b). Although the robots appear to speak to each other during the performance, the choreography is actually achieved using backstage communication *via* a blackboard architecture (Hayes-Roth, 1985). The NAO robots are bipedal, humanoid robots offering 25 degrees of freedom (Gouaillier et al., 2009). *Scéalextric* and *Scéalability* are used to generate and perform the stories that crowd-sourced judges will evaluate for spatial and gestural coherence. The test performances can combine spatial movements and gestures, or use spatial movements alone, or use gestures alone. In each case, spatial movements and gestures can be chosen coherently, on the basis of an interpretation of the plot, or incoherently. Incoherent spatial movements are chosen to be the opposite of what would be considered coherent in context; thus, if the coherent movement is to take a step forward, the incoherent alternative is to take a step backwards, and vice versa. For gestures, which have no clear opposite, the incoherent choice is a random choice amongst all available gestures.

Our first experiment has three conditions, and raters on the crowd-sourcing platform *Amazon Mechanical Turk* (*AMT*) are presented with stories reflecting one of these conditions. In the first condition, the performers use only spatial movements in the enactment, and not gestures. Those movements are always chosen to be coherent with the plot. In the second condition, the performers use only gestures, not spatial movements, where again those gestures are chosen to be coherent. In the third condition, performers use coherent spatial movements and coherent gestures in the same performance. Raters reflect their appreciation of the given performance, which they view as an online video, on a Human-Robot Interaction questionnaire. Their feedback then forms the basis of a between-subjects study. On average, and as shown in **Figure 6**, raters appear to appreciate coherent gestures more than coherent spatial movements, but appear to appreciate a coherent combination of both more than any one mode of physical expression.

For the experimental setup of three conditions, the null hypothesis ($H_0$) states that there are no differences in the appreciation of the performances with different types of body movements, i.e., the rating of the performance should be independent of the movement performed. Our alternative hypothesis ($H_1$) is that there are real differences in the appreciation of performances that use different body movements, i.e., the rating of a performance should be dependent on the type and coherence of the movements used. An analysis of variance (ANOVA) is conducted to determine whether there are any statistically significant differences between the means of the three conditions. Unless one or more of the distributions is highly skewed, or the variances are very different, the ANOVA is a reliable analytical measure. When we compare the variances of the distributions with a Levene test for homoscedasticity (Brown and Forsythe, 1974), no significant differences are found (Test-statistics = 2.207 with $p > 0.05$ to accept equal variances). Hence, we assume equal variance for all conditions.

Consequently, the analysis of variance reveals significant differences between the three conditions, with $p = 0.0019$ (Sum of squares = 38.686, F-values = 6.292). With the results of this ANOVA, we can reject the null hypothesis to argue that there is a significant dependence of the rating of the performance on the type and coherence of the movements used. We hypothesize that some movements are more appreciated than others, but since the ANOVA does not make any claims about individual differences and effects, we conducted an additional t-test. Given the significant differences between conditions of more-or-less equal variance, we applied a two-sided t-test to tease out the differences between the three conditions. Because the t-test only provides a $p$-value to signify in-between differences, we also calculated Cohen's d to measure an appropriate effect size for the comparison between the means of the three conditions. This t-test showed a significant difference between the spatial movement and combined movement conditions ($p = 0.002$ Bonferroni corrected). The spatial movement condition yielded a mean appreciation score of $\mu_{Spatial} = 3.728$ with a standard deviation of $\sigma_{Spatial} = 1.792$. The combined condition received an average appreciation rating of $\mu_{Combined} = 4.131$ with a standard deviation of $\mu_{Combined} = 1.762$. The measured effect favours the latter (Cohen's D = 0.227). Statistical tests have been conducted on the accumulated test construct (of all 14 items) and the results are visualized in **Figure 6**. More details are available in (Wicke and Veale, 2020a).

Within the Human-Robot Interaction questionnaire, each participant rates the robotic performance using an appreciation construct that comprises two parts of seven questions apiece. One part, which measures the perceived attractiveness of the performance, uses the *AttrakDiff* questionnaire of Hassenzahl et al. (2003). The other elicits quality ratings for the embodiment, e.g., as to whether the physical actions of the performers appear to be appropriate to the story. The scores in **Figure 7** represent mean average scores, which to say, mean scores for all fourteen questions averaged across all raters for the relevant conditions.

The stories generated using *Scéalextric* can be long and convoluted, with many sub-plots and secondary characters. This complexity tends to confound the analysis of embodiment choices, since it requires raters to watch long video performances. So, for the three conditions studied here, raters are shown extracts from longer performances that focus on specific events in a story that involve the kinds of movements we aim to evaluate. Three one-minute videos are shown to 40 raters for each condition (so 120 in total) on *Amazon Mechanical Turk*. Each one-minute video is an excerpt of an embodied performance with a narrative voice-over. Each rater is paid 0.40$ per video to fill out the questionnaire of 14 questions. Three additional gold-standard questions are also included, which allow us to detect disengaged raters who provide uniform or random responses. When we exclude these invalid responses, there are 32 valid responses for the Spatial Movement condition, 29 for the gesture condition and 33 for the combined movement condition, yielding a total of $N = 94$ valid responses. More details on this study can be found in (Wicke and Veale, 2020a).

The crowd-sourcing of raters on platforms such as *Amazon's Mechanical Turk (or AMT)* brings with it some clear advantages and disadvantages. AMT does not provide demographic information about its participants, so we did not seek out this information. While there are concerns about the demographic characteristics of AMT rater populations (Chandler and Shapiro, 2016), AMT can still provide a relatively diverse demography, especially if compared to other Web-based samples and to the average American campus sample (Buhrmester et al., 2016). A study of *AMT* workers by Michel et al. (2018) reports that the average age of task participants is 35.5 years (SD = 11.0), and that 58% of workers are female.

## 5.2 To Interpret or Not: Coherence Versus Incoherence

This first experiment concerns performances in which performers always make coherent choices. In a second experiment, we aim to show that audiences appreciate coherence more than incoherence – and thus appreciate interpretation over non-interpretation – by showing raters performances in which choices are made either coherently (using the interpretation framework) or incoherently (ignoring, or going against the interpretation framework). This second experiment creates performances that observe one of four conditions: using spatial movements alone (coherent), with no gestures; using spatial movements alone (incoherent), with no gestures; using gestures alone (coherent), with no spatial movements; and using gestures alone (incoherent), with no spatial movements. Once again, raters evaluate video performances from a given condition, and provide their ratings using the same Human-Robot Interaction questionnaire. A between-subjects study of their ratings, again collected *via Amazon Mechanical Turk* and that again incorporates gold-standard questions to weed out disengaged raters, yields the findings shown in **Figure 7**.

As before, 40 raters were recruited for each condition ($N = 40 \times 4 = 160$), and each was paid 0.40$ for completing the questionnaire after watching a 1-minute video. After filtering invalid responses, the four trials resulted in 29 valid responses for coherent gestures, 28 for incoherent gestures, 32 for coherent spatial movements and 29 for incoherent spatial movements ($N = 118$). Our findings suggest that audiences do appreciate coherent interpretation over the incoherent lack of interpretation when performers use physical actions to convey a story.

# 6 CONCLUSION

## 6.1 Frameworks for Storytelling

The relative strengths of the Performance and Interpretation frameworks underline the distinction between what is performed and how it is interpreted. Both systems are distinct, yet they must work together, because interpretation is based on performance, and the latter is shaped by what the system and its actors wish to convey. Performers must first interpret for themselves what they wish an audience to subsequently interpret from their actions. But this is hardly a novel concern. Within the theory and practice of acting, it is suggested that "Rather than playing an emotion, actors are advised to play the action and encode the emotion in the action through parameters, such as speed, intensity, shape, and direction." (El-Nasr, 2007). Human approaches to acting, such as that famously outlined in (Stanislavski, 2013), can thus inform our approach to the robotic performance of stories. Importantly, however, we must abstract away from the physical limitations and peculiar affordances of the actors themselves, or, in our case, of the specific robots that we employ. The modularity of our approach is a clear advantage in this regard.

Creativity by a producer always requires a corresponding (if perhaps lesser) creativity in the consumer if it is to be properly appreciated. In this paper we have necessarily focused on producer-side creativity, and said little about the consumer-side creativity that it necessitates in turn. This lopsided view is tenable in the short-term, for practical reasons, but it must be redressed eventually. Future work must thus address this imbalance, which is inherent to the creative equation in any performative context. Producers anticipate how consumers will react, while consumers model the intent of the producer. To adequately account for one side of the equation, we must also account for the other.

## 6.2 Discussion

The Softbank NAO robots that are used in our performances and crowd-sourced evaluations have none of the grace or agility exhibited by recent, bio-inspired machines, such as those of *Boston Dynamics* (Guizzo, 2019). Those robots are capable of animal-like movements and human-like poise, as recently shown in scripted robotic dances[2]. Nonetheless, we focus on a larger point here, one that is mostly independent of the robotic hardware that is used. To perform a story for an audience, performers must do more than follow a literal script to the letter. They must interpret the script, to actually fill the positions – spatially and otherwise – of the characters they are supposed to "inhabit." Interpretation requires an emotional understanding of the unfolding plot, so that actions can be chosen to coherently reflect that understanding.

To this end, our interpretation and performance frameworks employ representations and mechanisms that mediate between plot actions, character emotions, and a performer's movements and gestures on stage. Interpretation creates choice for a performer, motivating departures from the script when the scripted response seems inadequate in context. Moreover, interpretation guides performance, so that the robotic performers become part of a larger whole, in which relative position is as important as individual action.

One dimension of human emotional expression that is overlooked here is that of facial expression. We humans communicate with our looks as well as our words and gestures, as shown e.g., by the importance of non-manual features in sign language (Nguyen and Ranganath, 2012). We do not consider this dimension here because it is not the primary focus of our current work, not least because our robots lack the means to emote with their faces. Nonetheless, facial emotion is a dimension we must address in future work and in any addition to the current framework. To begin, we are presently considering the role of facial emotion and gestures in audience members as they watch a robotic performance. When viewers engage with a story, their expressions and gestures can subtly (and not so subtly) guide the interpretations of the robotic performers, perhaps warranting a comment in response, or even a dramatic plot change in mid-narrative. As outlined in (Wicke and Veale, 2021), the underlying stories can be generated as disjunctive trees rather than linear paths, and robots can elicit emotional feedback from users *via* a video feed, to determine which forks in the plot to follow.

We have defined a modular and extensible framework that can be adapted and reused by HRI researchers for different kinds of robotic performance. For example, a study on the perception of drone movements by Bevins and Duncan (2021) evaluates how participants respond to a selection of schematic flight paths. Since the identified movements are inherently schematic, e.g. Up-Down and Left-Right, we believe that our framework can help to categorize their results from a performative and an interpretative perspective. Other recent work suggests how we might extend our performance framework's taxonomy of motion types to include, as noted earlier, properties such as *naturalness*.

---

[2]https://www.youtube.com/watch?v=fn3KWM1kuAw

In this vein, Kitagawa et al. (2021) investigate how robots can most naturally move toward their goals, and show that common *rotate-while-move* and *rotate-then-move* strategies are inferior to their proposed set of human-inspired motion strategies.

A natural complement to the visual modality of spatial movement is sound. Three types of artificial sounds for robotic movements are explored in Robinson et al. (2021) in the context of a *Smooth Operator*. Their results indicate that robotic movements are interpreted differently, and perceived as more or less elegant, controlled or precise, when they are coupled with different sounds. These findings suggest that our interpretation framework might be applied to additional modalities, such as sound, to exploit additional channels of communication and augmented modes of meaning-making.

Of course, these modalities are usually bidirectional. The visual modality, for example, works in both directions: as the audience perceives the performers, the performers can also perceive their audience. As noted earlier, this allows audience members to use their own gestural and facial cues to communicate approval or disapproval to the actors as they enact their tales. This feedback – in the guise of a smile or a frown, a thumbs up or a thumbs down – allows performers to change tack and follow a different path through the narrative space when, as described in Wicke and Veale (2021), the underlying story contains branching points at which actors should seek explicit feedback from the audience. A video camera provides a visual feed that is analyzed for schematic gestures and facial emotions, and when such cues are present, the actors base their choice not on their own interpretation but on that of the audience.

## 6.3 The Nao Robot

When considering the limitations of this work, we must address our choice of robot, the Nao. Qualities that are desirable from an interpretative perspective may be undesirable from a performance perspective, and vice versa. For instance, a decision to link arousal with the energy and speed of the robot's actions must consider issues of unwanted noise (from the robot's gears) and balance (it may fall over if it reacts too dramatically). The latter also affects its use of space. As robots move closer together, to e.g., convey emotional closeness, their gestures must become more subtle, lest they accidentally strike one another in the execution of a sweeping motion. While our interpretation and performance frameworks might look different than they are had we chosen a different platform, we are confident that these modular and extensible frameworks can grow to accommodate other choices in the future, either by us or by others.

Because the Nao platform has been used in a variety of related research [e.g., Gelin et al. (2010), Pelachaud et al. (2010), Ham et al. (2011), LaViers and Egerstedt (2012), and Wicke and Veale (2018a)], this speaks well to the reproducability of our approach. Despite its limitations, the Nao currently suits our needs, not least because it has 25 degrees of freedom and the ability to move its limbs independently of each other. The robot's fixed facial expression is certainly a limitation, one that prevents us from conveying emotion with facial cues, yet this also helps us to avoid unwanted bias in our user studies. It also means that we need not worry that the robot's manual gestures will occlude its facial expressions at key points. Other limitations can be addressed in a more-or-less satisfactory fashion. For instance, the Nao cannot turn on the spot, but turning can be implemented as a composition of spatial and rotational movements. So, although the movements of our performance framework are shaped in large part by the abilities of our robots, they are not wholly determined by their limitations. A comparison with other robot platforms would undoubtedly be useful and revealing, but it is beyond the scope of this current paper.

## 6.4 Current Thoughts, Future Directions

The interpretation and performance frameworks support both fine-grained and gross-level insights into the unfolding narrative. For instance, we have seen that aggregate assessments of valence – for a given role of a specific action at a particular point in the plot – allow for aggregate judgments about characters and their changing feelings. These gross judgments, which reveal positive or negative shifts in a character's overall feelings, can suggest equally reductive actions for robot performers to execute on stage, such as moving closer to, or further away from, other performers playing other characters. In this way, gross interpretations support powerful spatial metaphors that are equally summative and equally persistent. We have largely focused here on the semantics of gross spatial actions, but the literature provides a formal basis for the more fine-grained forms of expression that we will pursue in future work, such as those from the domain of dance (LaViers et al., 2014; Bacula and LaViers, 2020).

But fine-grained insights are also supported by the framework, which is to say, insights based on movements along a single emotional dimension. Spatial movement to and fro, of the kind evaluated in the previous section, are reductive and general. But metaphors that allow a performer to construe on action as another in a given context, such as by construing an *insult* as an *attack*, or an act of *praise* as an act of *worship*, are more more specific. They work at the conceptual level of plot action, and do more than suggest an embodied response. Rather, they increase the range of choices available to a performer because they operate at a deeper and more specific level of interpretation.

We humans reach for a metaphor when we want to broaden our palette of expressive options, and so too can our robot performers. But metaphor it itself just one choice that leads to others. Irony is another. A performer can, for example, choose to react ironically to a script directive. Suppose character A is expected to show fealty to character B, and the story so far firmly establishes this expectation in the minds of the audience (and in the view of the interpretation framework). Irony is always a matter of critiquing a failed expectation, by acting as though it has not failed while clearly showing that it has. It is the ultimate creative choice. Suppose now that the plot calls for A to *rebel against*, or *stand up to*, or to *break with* B. When the interpretation framework compares the emotions established by previous actions with those stirred by this new action, it recognizes a rift that should, if it is large enough, influence how the performers react. The robot portraying character A might thus act out an action more in line with the expected emotions, such as bowing down to B, while speaking the dialogue associated with the current action, such as "I've had enough of you!". The bifurcation of irony, of expectation vs. reality, easily maps onto the parallel modalities of speech and physical action, so that a performer can indeed follow both branches at once.

Although we have not examined or evaluated irony here, we mention it now to show that robotic performances of complex semiotic structures, such as stories, open many avenues for an interpretative performer, at both the conceptual and the expressive levels. These choices, which include construal mechanisms such as metaphor and irony and more besides, open more choices in turn, if a performer has the wit to perceive and exploit them. As such, it is fair to say that we have barely scratched the surface of what an interpretative approach to embodied performance can yet bring to domains such as story-telling. As we go deeper, we may need to use a richer model of the gestures and motions that realize the embodiment, such as by drawing on insights and representations from the world of dance, where more nuanced actions – and more nuanced notations – necessarily hold sway.

## DATA AVAILABILITY STATEMENT

The datasets presented in this study can be found in online repositories. The names of the repository/repositories and accession number(s) can be found below: Movement and Gesture Repository for Robots and Humans (at OSF) https://osf.io/e5bn2/?view_only=2e30ee7e715342d59c371b5d30c014e0.

## ETHICS STATEMENT

The studies involving human participants were reviewed and approved by the Ethics Committee of the authors' university, University College Dublin, Ireland, under protocol number UCD HREC-LS, Ref. No.: LS-E-19-125-Wicke-Veale. The study received exemption from ethics approval. The patients/participants provided their written informed consent to participate in this study.

## AUTHOR CONTRIBUTIONS

Both authors contributed to conception and design of the research. The first implemented the robotic frameworks and prepared an initial draft of the manuscript. The second implemented the story-generation framework. Each author contributed equally to all sections of the manuscript, to revising the manuscript, and to preparing and submitting the current version.

## SUPPLEMENTARY MATERIAL

The Supplementary Material for this article can be found online at: https://www.frontiersin.org/articles/10.3389/frobt.2021.662182/full#supplementary-material